\documentclass{article}
\usepackage{preprint_layout,times}
\usepackage{amsmath,amssymb,amsthm,graphicx,booktabs,array,multirow,longtable}
\usepackage{hyperref,url}
\usepackage{placeins}
\hypersetup{colorlinks=true,linkcolor=black,citecolor=black,urlcolor=black,pdftitle={From Soft Targets to Reward Signals: How Assignment and Reward Objectives Interact},pdfauthor={Jiangtao Lin, Bangyang Wei, Siyi Liu, Yihang Ding, Yuhan Dong}}
\renewcommand{\headrulewidth}{0pt}
\pdftrailerid{}
\newtheorem{proposition}{Proposition}
\newcommand{\E}{\mathbb{E}}
\newcommand{\R}{\mathcal{R}}
\newcommand{\A}{\mathcal{A}}
\newcommand{\U}{\mathcal{U}}
\newcommand{\tabfont}{\normalsize}
\title{From Soft Targets to Reward Signals: How\\Assignment and Reward Objectives Interact}
\author{Jiangtao Lin\textsuperscript{1}, Bangyang Wei\textsuperscript{2}, Siyi Liu\textsuperscript{1,4}\\
\textbf{Yihang Ding\textsuperscript{1,3}, Yuhan Dong\textsuperscript{1,*}}\\[0.4em]
\textnormal{\small \textsuperscript{1}Tsinghua Shenzhen International Graduate School, Tsinghua University}\\
\textnormal{\small \textsuperscript{2}School of Vehicle and Mobility, Tsinghua University}\\
\textnormal{\small \textsuperscript{3}SZ DJI Technology Co., Ltd.}\\
\textnormal{\small \textsuperscript{4}Tencent Holdings Limited}\\[0.3em]
\textnormal{\small \textsuperscript{*}Corresponding author: \href{mailto:dongyuhan@sz.tsinghua.edu.cn}{dongyuhan@sz.tsinghua.edu.cn}}}

\begin{document}
\maketitle
\begin{abstract}
Soft preference targets specify supervision strength, and reward objectives convert that strength into learned reward signals. A central design question remains: how does assigning a fixed set of preference strengths to different response pairs change the rewards produced by different objectives? We introduce assignment geometry to study this interaction. Mean-matched smoothing controls target dispersion, while within-stratum reassignment changes correspondence and preserves the complete target distribution. Across five reward objectives, intact correspondence retains the largest clean preference margins among the compared soft targets within a common accuracy-equivalence budget. Attenuation orderings change with the reward objective, revealing different responses to the same target assignments. Independent reassignments and a related source construction reproduce the retention direction. An attenuation-retention profile compares these combinations through margin magnitude, edit response, and accuracy. Against independently calibrated scaling, APLOT uniform targets deliver additional attenuation on both aggregate and presentation edits. These findings establish a joint design space in which target placement and reward objective shape reward properties beyond preference accuracy.
\end{abstract}

\section{Introduction}
Preference learning turns comparisons between language-model responses into reward signals. A clear preference and a near tie carry different information; soft targets express this distinction by assigning a desired preference probability to each pair. Relative-quality targets, ordinal feedback, and adaptive margins have made this supervision increasingly expressive \citep{kim2024mmpo,liu2025ordinalfeedback,li2025aplot}. Each target enters learning through a reward objective, connecting supervision strength to the margins and input responses that the trained scorer produces.

These two design choices draw on established foundations. Instance-specific smoothing connects confidence diversity and assignment to classification learning \citep{zhang2020instance}. Reward objectives shape pairwise updates through adaptive margins, representation normalization, reward regularization, and context dependence \citep{li2025aplot,xie2026normbt,hong2025bsr,liu2026darm}. These advances leave unresolved how the same target assignments translate into clean margins and edit responses across reward objectives. We therefore ask: \emph{which assignment effects persist across objectives under a common supervision budget, and which depend on the learning rule?}

Our experiments show that the reward objective changes the attenuation ordering of target placements. Uniform targets yield stronger aggregate edit attenuation than intact targets under BT and APLOT; NormBT reverses this ordering. Among the soft placements, intact correspondence retains the largest clean preference margins across all five evaluated objectives at comparable accuracy. Thus choosing a target placement also requires choosing the objective through which it will act. Pair accuracy alone leaves these differences in learned reward signals unresolved.

\begin{figure}[t]
\centering
\includegraphics[width=\linewidth]{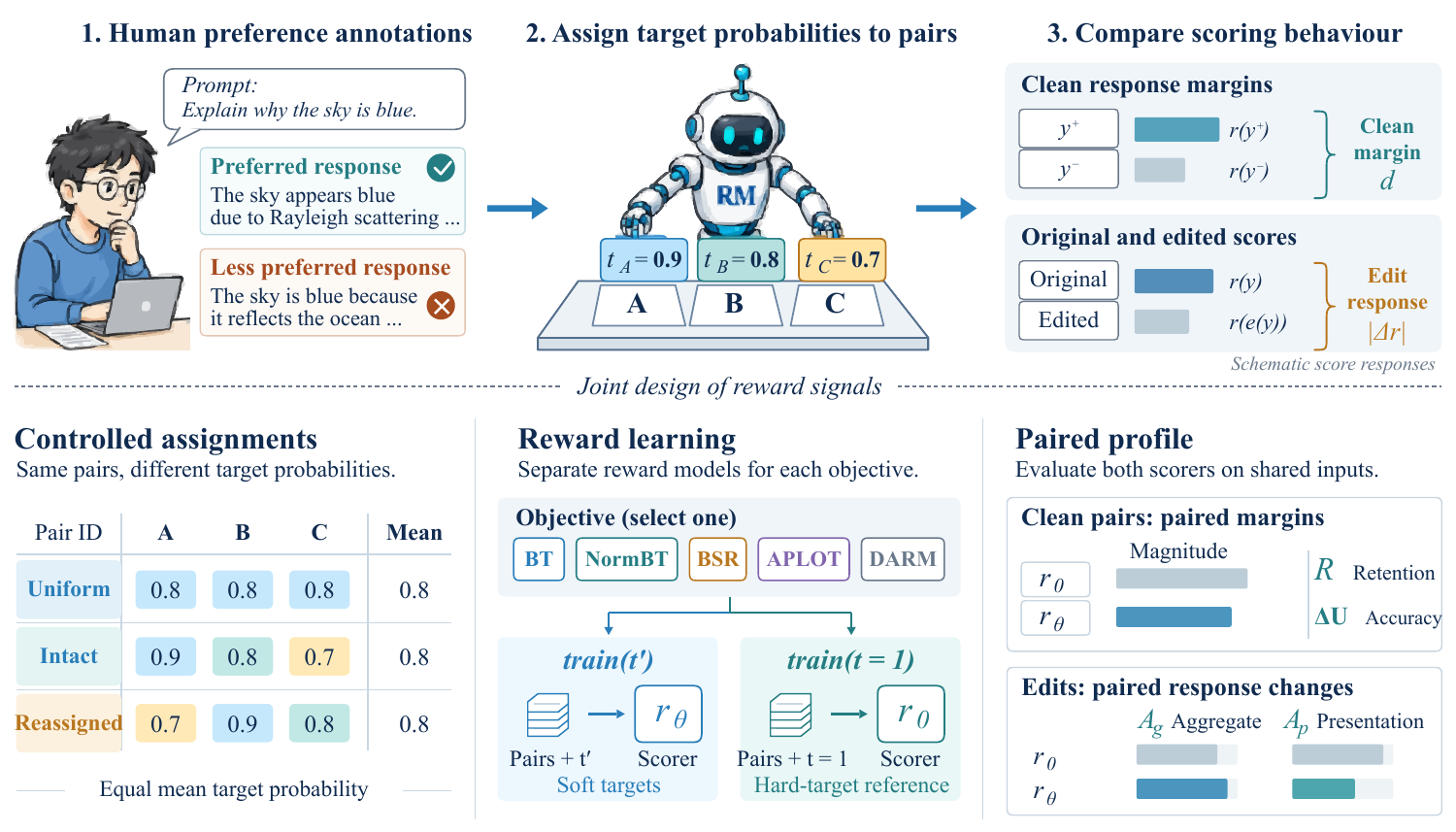}
\caption{\textbf{Target placement and reward objective jointly shape the learned profile.} The three placements enter each of five reward objectives. Each objective trains a soft-target scorer and a hard-target raw reference. Both scorers evaluate the same clean pairs and edits, yielding retention $R$, accuracy change $\Delta U$, and aggregate and presentation attenuation $A_g,A_p$. Target probabilities illustrate the assignments; the prompt and bar lengths are schematic.}
\label{fig:framework}
\end{figure}

To study this interaction, we introduce \emph{assignment geometry} (Figure~\ref{fig:framework}). A mean-matched control removes within-stratum dispersion, while reassignment keeps response pairs fixed and preserves the complete within-stratum target distribution. Crossing these controlled placements with reward objectives separates the contribution of target values from their attachment to training pairs. An \emph{attenuation-retention profile} describes the resulting edit responses and clean preference-margin magnitude. Independently calibrated scaling then identifies the additional attenuation delivered by each trained combination.

We evaluate this design across five reward objectives, independent distribution-preserving reassignments, and a related source construction, then compare the learned responses with calibrated scaling. Our main contributions are as follows:
\begin{itemize}
\item We introduce \textbf{assignment geometry}, making dispersion and pair attachment separately controllable through mean-matched smoothing and distribution-preserving reassignment. Crossing these placements with reward objectives isolates how each learning rule expresses the same supervision.
\item We reveal a \textbf{shared retention effect and objective-dependent attenuation}: intact targets retain the largest clean margins across five objectives within a common accuracy-equivalence budget, while the objective changes the attenuation ordering of target placements.
\item We develop an \textbf{attenuation-retention profile} for joint target and objective design. Independently calibrated scaling identifies extra attenuation delivered by trained combinations, with APLOT uniform targets producing additional response reduction on both edit banks.
\end{itemize}

\section{Related Work}
\label{sec:related}
\paragraph{Soft supervision and confidence allocation.}
Distillation transfers predictive distributions \citep{hinton2015distilling}, while label smoothing changes calibration and represented class relationships \citep{muller2019smoothing}. Permuted non-target predictions \citep{furlanello2018bornagain} and confidence diversity with matched-average and randomized assignments \citep{zhang2020instance} establish allocation as a learning variable. A statistical account relates distillation to probability-estimation quality and objective variance \citep{menon2021statistical}. For reward learning, this perspective raises a further question: how does an allocation shape the reward signal under different training objectives? We study this question by linking controlled target assignments to clean margins and edit responses.

\paragraph{Graded preference supervision.}
In language-model alignment, comparison feedback trains rewards and policies \citep{ouyang2022instructgpt}. MMPO uses relative-quality margins to construct soft probabilities \citep{kim2024mmpo}; ordinal-feedback learning incorporates preference strength and crowd judgments \citep{liu2025ordinalfeedback}. Selective smoothing and adaptive margins enrich reward supervision \citep{wang2024secretsrlhfpartii}. DPO provides a policy-level preference objective \citep{rafailov2023dpo}, with conservative DPO introducing constant target smoothing \citep{mitchell2023cdpo}. These developments provide ways to encode preference strength. Given a source of soft targets, separating the effects of their values and their attachment to pairs requires matched comparisons. Our construction controls dispersion at a fixed mean correction mass; reassignment changes pair attachment while preserving the full within-stratum target distribution.

\paragraph{Reward objectives and update structure.}
Target placement acts through the objective used to learn a reward, whose behavior matters for overoptimization \citep{gao2023overoptimization,rafailov2024directoveroptimization}. APLOT adapts margins using semantic similarity, reward differences, and optimal transport \citep{li2025aplot}. NormBT normalizes representation-distance effects \citep{xie2026normbt}; BSR regulates batch reward sums \citep{hong2025bsr}; DARM strengthens preference-context dependence \citep{liu2026darm}. These mechanisms change how assigned targets enter learning updates, raising a shared question: which assignment effects persist across objectives, and which depend on the learning rule? Our matched comparisons show higher clean-margin retention with intact correspondence across these objectives, while attenuation orderings depend on the learning rule.

\paragraph{Reward measurement and controls.}
Comparing these learned rewards requires separating preference selection, reward magnitude, and edit response. RewardBench evaluates preference selection across tasks \citep{lambert2025rewardbench}, and RM-Bench separates subtle content differences from stylistic variation \citep{liu2025rmbench}. Calibration complements accuracy with probability quality \citep{guo2017calibration}, while control tasks tie measurement to explicit interventions \citep{hewitt2019controltasks}. For assignment comparisons, reward scale adds a specific ambiguity: positive rescaling changes margins and edit responses while leaving preference accuracy fixed. We therefore report these properties jointly in a paired profile and use independently calibrated scaling to measure edit attenuation beyond matched scalar controls.

\section{Assignment Geometry}
\label{sec:geometry}
\subsection{A controlled space of soft targets}
Our construction fixes the average amount of softening while controlling how target values vary and which pairs receive them. Let $\mathcal D=\{(x_i,y_i^+,y_i^-,t_i)\}_{i=1}^n$ contain prompts, preferred and dispreferred responses, and targets $t_i\in[1/2,1]$. Define correction mass $m_i=1-t_i$ and reward margin $d_{\theta,i}=r_\theta(x_i,y_i^+)-r_\theta(x_i,y_i^-)$. With sigmoid $\sigma(d)=(1+e^{-d})^{-1}$, the soft Bradley--Terry loss is
\begin{equation}
\ell_{\rm BT}(d,t)=-t\log\sigma(d)-(1-t)\log[1-\sigma(d)].
\label{eq:bt}
\end{equation}
This loss makes the desired preference probability explicit. Assignment geometry is defined relative to a source vector $m$ and a stratum partition $g(i)$. A selected reward objective $\mathcal L_j$ receives these assigned targets, with its normalization, adaptive margins, regularization, or context terms determining their role in training.

\emph{Correction mass} $M=n^{-1}\sum_i m_i$ is the average departure from hard labels. For a stratum $g$ containing $n_g$ pairs, \emph{dispersion} describes variation around its mean $\bar m_g$. \emph{Correspondence} records which pair receives each value. We vary dispersion and correspondence through
\begin{equation}
m_i^{(\lambda,p)}=\bar m_{g(i)}+\lambda\bigl(m_{\pi_{g,p}(i)}-\bar m_{g(i)}\bigr),
\qquad t_i^{(\lambda,p)}=1-m_i^{(\lambda,p)},
\label{eq:geometry}
\end{equation}
where $\lambda\in[0,1]$ scales dispersion and $\pi_{g,p}$ reassigns a fraction $p$ of rows within each stratum. Targets equal the stratum mean at $\lambda=0$ and recover the source assignment at $(\lambda,p)=(1,0)$. At fixed $\lambda$, reassignment preserves the complete multiset. Every configuration shares the stratum mean correction mass, and its within-stratum standard deviation equals $\lambda$ times the corresponding source standard deviation.

These controls preserve different information about the supervision. Uniform keeps each stratum's mean while replacing pair-specific variation with a common strength. Reassigned retains every target value, so all quantiles and moments of its distribution agree with Intact; only the attachment to fixed response pairs changes. The first comparison asks how dispersion shapes learning. The second asks how pairing the same strengths with different examples changes the reward. Applying both comparisons within each objective separates these two contributions to the learned signal.

Table~\ref{tab:geometry} summarizes the three placements. Keeping separate stratum means preserves source-specific supervision budgets across the mixed preference data. We apply these interventions to a representation-based source constructor, denoted LCC. It produces out-of-fold anchor probabilities from a model fit with frozen representations, surface features, and source interactions, then mixes the anchors with hard preferences. A related length-based variant changes the explicit surface features while retaining the representation construction and mixing coefficient (Appendix~\ref{app:protocol}).

\begin{table}[t]
\centering\tabfont\setlength{\tabcolsep}{7pt}\renewcommand{\arraystretch}{1.12}
\caption{\textbf{Three placements of the same correction budget.} Every configuration preserves the stratum mean $\bar m_g$ and fixed response pairs. $\sigma_g$ is the source within-stratum standard deviation.}
\label{tab:geometry}
\begin{tabular}{lcccc}
\toprule
Target & $(\lambda,p)$ & Assigned mass & Dispersion & Pair attachment \\
\midrule
Uniform & $(0,0)$ & $\bar m_g$ & $0$ & Constant within stratum \\
Intact & $(1,0)$ & $m_i$ & $\sigma_g$ & Source correspondence \\
Reassigned & $(1,1)$ & $m_{\pi_g(i)}$ & $\sigma_g$ & Permuted within stratum \\
\bottomrule
\end{tabular}
\end{table}

\subsection{Target logits and shared updates}
The construction separates two routes from targets to learning. Dispersion changes the distribution of desired margins: for independent logits with $0<m_i<1/2$, Equation~\ref{eq:bt} has minimizer $d_i^*=\log[(1-m_i)/m_i]$.
\begin{proposition}[Matched mass in an independent-logit model]
For a stratum with $0<m_i<1/2$,
\begin{equation}
\frac{1}{n_g}\sum_{i\in g}\log\frac{1-m_i}{m_i}
\ \geq\ \log\frac{1-\bar m_g}{\bar m_g},
\label{eq:jensen}
\end{equation}
with equality exactly when its correction masses are constant.
\end{proposition}
The positive second derivative $(1-2m)/(m^2(1-m)^2)$ gives the result by Jensen's inequality. Uniform allocation therefore minimizes the average optimal logit at fixed mean mass, identifying a geometric source of contraction. Appendix~\ref{app:proof} develops the interpolation relation. Reassignment preserves the multiset of these independent optimal logits, including their mean. Its effect on a shared scorer must therefore be studied through the association between targets and the pairs that receive them.

Correspondence supplies the second route, linking each target to a pair-specific update. For the empirical BT loss $L(t)=n^{-1}\sum_i\ell_{\rm BT}(d_{\theta,i},t_i)$, reassignment from $t$ to $t'$ at fixed parameters gives
\begin{equation}
\nabla_\theta L(t')-\nabla_\theta L(t)
=\frac{1}{n}\sum_i(t_i-t'_i)\nabla_\theta d_{\theta,i}.
\label{eq:gradient}
\end{equation}
Target differences sum to zero within each stratum, while their products with margin Jacobians depend on the receiving pairs. If all margin Jacobians within a stratum coincide, its contribution to Equation~\ref{eq:gradient} cancels. Pair-dependent Jacobians allow the same target values to direct different shared updates. Dispersion controls deviation size, correspondence places deviations on particular Jacobians, and reward objectives transform their contributions through normalization or additional losses. These two routes motivate measuring each objective's response to both controlled changes in supervision.

\section{Measuring the Learned Profile}
\label{sec:profile}
\paragraph{Clean-margin retention and accuracy.}
To connect these updates to learned rewards, we measure margin retention, accuracy, and edit response against a hard-target raw control $r_0$ for each objective and seed. Its row-average absolute clean margin defines $s_0$. Let $\E_P$ average outcomes within prompts and then prompts equally. Define
\begin{align}
\R(\theta)&=1+\frac{\E_P[|d_\theta|-|d_0|]}{s_0},\label{eq:retention}\\
\Delta\U(\theta)&=\E_P[\mathbf1\{d_\theta>0\}-\mathbf1\{d_0>0\}].\label{eq:utility}
\end{align}
Raw retention equals one. $\R$ measures retained clean preference-margin magnitude, and $\Delta\U$ measures accuracy change on the recorded preferences (Appendix~\ref{app:margin-diagnostics}).

\paragraph{Edit attenuation.}
For an edit $e$ of response $y$, let $b_\theta(e)=r_\theta(x,e(y))-r_\theta(x,y)$. Define
\begin{equation}
\A_{\mathcal E}(\theta)=\frac{\E_{P,e\in\mathcal E}[|b_0(e)|-|b_\theta(e)|]}{s_0}.
\label{eq:attenuation}
\end{equation}
Positive values indicate a smaller absolute edit response. We distinguish an aggregate bank $\mathcal E_g$ of length, sentiment, and agreement edits from a presentation bank $\mathcal E_p$ of exactly reversible formatting operations. Separate evaluation on clean pairs and both edit banks captures different reward properties; the shared raw scale supports paired target contrasts.

\paragraph{Assignment contrasts.}
Let $u=(0,0)$ denote stratum-uniform targets, $i=(1,0)$ intact targets, and $s=(1,1)$ full reassignment. Our endpoint contrasts are
\begin{equation}
\Delta_\lambda=\A_g(u)-\A_g(i),\quad
\Delta_{p,A}=\A_g(s)-\A_g(i),\quad
\Delta_{p,R}=\R(i)-\R(s).
\label{eq:contrasts}
\end{equation}
These comparisons isolate dispersion at intact correspondence and correspondence at full dispersion. Reporting $\A_{\mathcal E}$ and $\R$ jointly yields the attenuation-retention profile. Together with accuracy, it shows how a placement changes edit response and clean-margin magnitude under a shared supervision budget.

\paragraph{A calibrated scaling reference.}
To compare placements at similar signal strength, we use positive scaling, which preserves a scorer's pair ordering and changes both profile coordinates. For each seed, we fit $c_\theta=\E_{P\in\mathcal C}|d_\theta|/\E_{P\in\mathcal C}|d_0|$ on clean calibration prompts $\mathcal C$ and use $c_\theta r_0$ as the reference. This matches the candidate's average absolute clean margin on calibration data. Evaluation on disjoint prompts then measures extra edit attenuation over the scaled reference and how closely the margin match carries to new pairs. Because calibration uses clean pairs alone, held-out edit responses provide a separate comparison of the trained and scaled scorers. Appendix~\ref{app:scalar} gives the full analysis.

\section{Experimental Design}
\label{sec:experiments}
\paragraph{Matched reward learning.}
We compare BT with objectives that modify normalization, reward regularization, adaptive margins, and context dependence: NormBT, BSR, APLOT, and DARM. Each objective receives uniform, intact, and reassigned targets alongside its own raw control. A common DeBERTa-v3-base backbone and fixed training pairs isolate how these learning rules express the same assignments \citep{he2023debertav3}. Four configurations at three seeds for each of five objectives give 60 reward models. Training uses 99,926 pairs and 7,697 validation pairs from sources including UltraFeedback \citep{cui2024ultrafeedback} and HelpSteer, with effective batch 32, learning rate $2\times10^{-5}$, and one epoch. APLOT follows its released reward-distance cost \citep{aplotcode}. Appendix~\ref{app:protocol} details initialization, objective mechanisms, and training conditions.

\paragraph{Assignment and source comparisons.}
To examine the retention effect across different mappings of the same target values, we cross three independent within-stratum reassignments with seeds $7,17,29$. The related source construction supplies uniform, intact, and reassigned targets at those seeds, testing whether the effect persists when the source values change. Both studies use paired raw references; the assignment study shares one intact reference per seed across its three mappings. A nine-edit study uses these same raw/intact references to characterize attenuation across presentation styles. Appendix~\ref{app:protocol} specifies the reference models and populations for each comparison.

\paragraph{Evaluation and comparison.}
All primary reward comparisons use 22,561 clean pairs with prompts disjoint from training, validation, and the existing test split, plus 21,360 aggregate-edit and 10,680 presentation-edit observations. Scaling uses separate calibration and evaluation prompts. We average training seeds equally and compare targets on shared prompts, giving each placement the same evaluation conditions. Independent-assignment results also average the three reassignments. Main tables and figures report measured profiles and paired gains; confidence intervals, multiple-comparison procedures, accuracy-equivalence tests, and seed sensitivity appear in Appendices~\ref{app:inference}--\ref{app:seed-sensitivity}. Appendix~\ref{app:scalar} reports the calibrated scaling comparisons.

\section{Results}
\label{sec:results}
\subsection{Correspondence retains clean preference margins}
\begin{figure}[t]
\centering
\includegraphics[width=\linewidth]{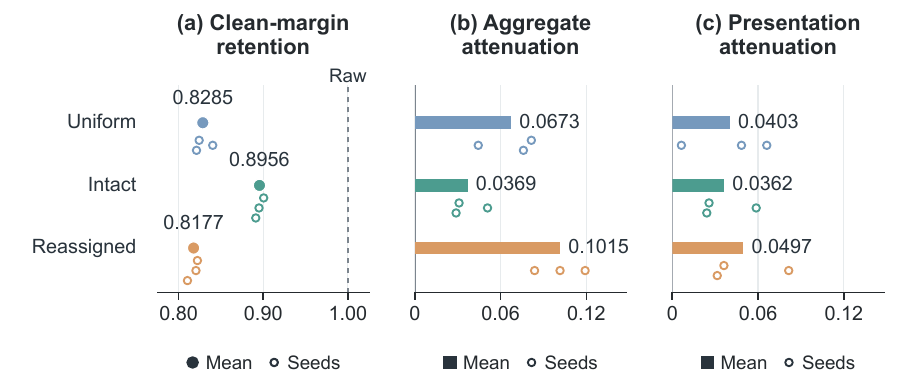}
\caption{\textbf{A common accuracy budget accommodates distinct reward profiles.} Rows align BT placements. Filled points and bars show means; open circles show individual seeds, offset vertically for visibility. Raw defines unit retention and zero attenuation.}
\label{fig:bt-profile}
\end{figure}

At comparable pair accuracy, target placement produces distinct combinations of clean-margin retention and edit attenuation. Intact correspondence retains the largest margins among the compared BT soft targets (Table~\ref{tab:bt-profile-main}). Its retention is $0.8956$, compared with $0.8285$ for uniform targets and $0.8177$ for reassignment. The intact advantage over reassignment is therefore $0.0779$. Reassignment simultaneously produces the strongest aggregate attenuation (Figure~\ref{fig:bt-profile}). All three geometry contrasts agree in direction across seeds, showing how placement controls the balance between edit attenuation and retained reward magnitude.

The paired coordinates explain the ordering more fully. Relative to intact targets, reassignment produces a larger increase in aggregate attenuation than in presentation attenuation, alongside lower clean-margin retention. Each coordinate compares the same candidate with the same raw reference. The resulting profile exposes how the effect is distributed across clean preference pairs, broad response perturbations, and reversible formatting changes. A target's position therefore depends on which reward property the comparison emphasizes.

\begin{table}[t]
\centering\tabfont\setlength{\tabcolsep}{7pt}
\caption{\textbf{Distinct BT reward profiles within a common accuracy budget.} Three-seed means on the strict population. All three targets satisfy accuracy equivalence to raw within one percentage point. Bold marks the largest soft-target retention.}
\label{tab:bt-profile-main}
\begin{tabular}{lrrrr}
\toprule
& \multicolumn{1}{c}{Accuracy change} & \multicolumn{2}{c}{Edit attenuation} & \multicolumn{1}{c}{Retention} \\
\cmidrule(lr){3-4}
Target & Difference (pp) & Aggregate & Presentation & Clean margin \\
\midrule
Uniform & $-0.091$ & $0.0673$ & $0.0403$ & $0.8285$ \\
Intact & $+0.003$ & $0.0369$ & $0.0362$ & $\mathbf{0.8956}$ \\
Reassigned & $+0.067$ & $0.1015$ & $0.0497$ & $0.8177$ \\
\bottomrule
\end{tabular}
\end{table}

Preserving correspondence also produces a consistent retention advantage across objectives (Table~\ref{tab:objectives}). Intact targets achieve the largest retention in each row and exceed reassignment by $0.0639$--$0.0946$. The effect holds across adaptive margins, normalization, regularization, and context-dependent training: these objectives all retain more clean-margin magnitude when source values remain attached to their original pairs. Because reassignment preserves the complete target distribution, this comparison identifies pair attachment as a shared determinant of reward magnitude across the five learning rules.

\begin{table}[t]
\centering\tabfont\setlength{\tabcolsep}{8pt}\renewcommand{\arraystretch}{1.12}
\caption{\textbf{Intact correspondence retains larger margins across reward objectives.} Values are three-seed means; gains compare intact with reassigned targets. Bold marks the largest retention among each objective's soft targets.}
\label{tab:objectives}
\begin{tabular}{lrrrr}
\toprule
& \multicolumn{3}{c}{Clean-margin retention} & \multicolumn{1}{c}{Correspondence} \\
\cmidrule(lr){2-4}\cmidrule(lr){5-5}
Objective & Uniform & Intact & Reassigned & Gain \\
\midrule
BT & $0.8285$ & $\mathbf{0.8956}$ & $0.8177$ & $+0.0779$ \\
NormBT & $0.8671$ & $\mathbf{0.9214}$ & $0.8575$ & $+0.0639$ \\
BSR & $0.8420$ & $\mathbf{0.9260}$ & $0.8314$ & $+0.0946$ \\
APLOT & $0.8596$ & $\mathbf{0.8940}$ & $0.8266$ & $+0.0674$ \\
DARM & $0.7776$ & $\mathbf{0.9105}$ & $0.8322$ & $+0.0783$ \\
\bottomrule
\end{tabular}
\end{table}

All fifteen reward-model configurations satisfy raw-relative accuracy equivalence within one percentage point (Appendix~\ref{app:utility}). Within this common budget, the profiles distinguish target placements by their retained margins and edit responses. We next examine how the reward objective changes the attenuation ordering of these placements.

\subsection{Reward objectives reshape attenuation}
\begin{figure}[t]
\centering
\includegraphics[width=\linewidth]{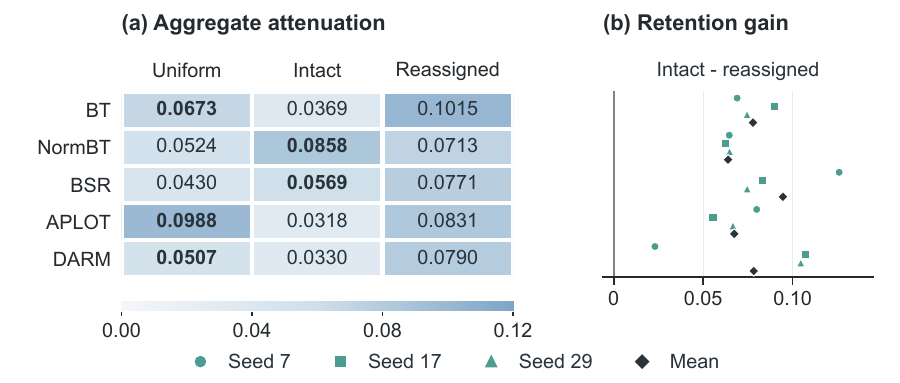}
\caption{\textbf{Reward objectives change attenuation orderings while preserving the retention direction.} Left: mean aggregate attenuation; bold marks the larger value among uniform and intact targets in each row. Right: intact-minus-reassigned retention for each seed and their mean. All values use objective-specific raw references.}
\label{fig:objectives}
\end{figure}

The shared retention direction accompanies distinct attenuation orderings (Figure~\ref{fig:objectives}). BT and APLOT give stronger aggregate attenuation to uniform and reassigned targets. Under NormBT, intact dispersion yields stronger attenuation than uniform smoothing. Thus the same target placements acquire different response properties under different reward objectives.

Direct comparisons establish this objective dependence. Uniform's attenuation gain over intact is $0.0304$ under BT, $0.0670$ under APLOT, and $-0.0334$ under NormBT. The difference between APLOT and NormBT is $0.1004$ in the direct interaction comparison (Appendix~\ref{app:interactions}). Each within-objective contrast uses the objective's shared raw scale; the interaction compares those assignment effects on common evaluation prompts.

For attenuation-oriented design, this changes which placement merits consideration. Uniform targets lead the intact configuration under BT and APLOT, while intact targets lead under NormBT. The source targets and assignment operations are held fixed across these learning rules. Transferring a smoothing choice between objectives therefore requires measuring the receiving objective's response. Alongside this changing attenuation order, preserving correspondence retains more clean-margin magnitude in every evaluated objective, providing a common reference for comparing their different profiles.

The comparison connects the two learning routes in Section~\ref{sec:geometry}. Equation~\ref{eq:jensen} characterizes independent target logits, and Equation~\ref{eq:gradient} locates correspondence in pair-dependent updates. The experiments measure the profiles produced after shared-model training, linking the same controlled assignments to a common retention direction and changing attenuation orderings.

\subsection{Correspondence retains clean margins across assignments and sources}
The common retention direction across objectives raises a further question: does it persist across different distribution-preserving reassignments? Table~\ref{tab:replication} compares intact targets with three independent reassignments. Intact correspondence gains $0.0761$ in mean retention, with all nine realization--seed combinations agreeing in direction. Gains for individual reassignments range from $0.0714$ to $0.0813$. The retention effect therefore persists across independently chosen receiving mappings of the same target values.

Varying source construction tests a different part of this relationship. The related length-based variant again gives intact correspondence higher retention and reassignment stronger attenuation, while all three configurations satisfy accuracy equivalence to raw. The independent mappings hold the source values fixed and vary their receivers; the source comparison changes the values before applying the same matched interventions. Together, these comparisons locate the retention pattern in the association between supervision strengths and training pairs across both kinds of variation.

\begin{table}[t]
\centering\tabfont\setlength{\tabcolsep}{6pt}\renewcommand{\arraystretch}{1.12}
\caption{\textbf{Correspondence retains clean margins across mappings and source constructions.} Each study uses paired references and averages three seeds; independent assignments also average three reassignments. Gains are intact minus reassigned for retention, and reassigned minus intact for attenuation.}
\label{tab:replication}
\begin{tabular}{lrrrr}
\toprule
& \multicolumn{2}{c}{Clean-margin retention} & \multicolumn{2}{c}{Paired gain} \\
\cmidrule(lr){2-3}\cmidrule(lr){4-5}
Study & Intact & Reassigned & Retention & Attenuation \\
\midrule
Independent assignments & $0.9097$ & $0.8336$ & $+0.0761$ & $+0.0355$ \\
Related source variant & $0.9133$ & $0.8248$ & $+0.0885$ & $+0.0386$ \\
\bottomrule
\end{tabular}
\end{table}

\subsection{Joint design of target placement and reward objectives}
The objective-dependent orderings make target placement and objective choice coupled decisions. The profile compares the resulting combinations through retained margin, edit response, and accuracy. Calibrated scaling adds a practical reference: it quantifies the extra attenuation supplied by training at a matched calibration strength.

\begin{figure}[t]
\centering
\includegraphics[width=\linewidth]{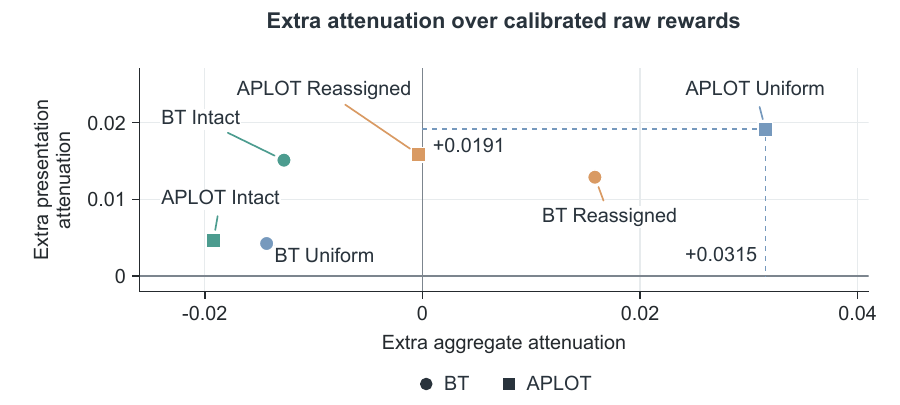}
\caption{\textbf{APLOT uniform targets deliver extra attenuation on both edit banks.} Each point shows a placement's mean aggregate and presentation residuals from its own calibrated raw reference. Circles denote BT and squares APLOT. Projections mark APLOT uniform's gains.}
\label{fig:scaling}
\end{figure}

APLOT uniform targets deliver attenuation beyond calibrated scaling on both edit banks (Figure~\ref{fig:scaling}). At retention $0.8596$, they achieve extra aggregate attenuation of $0.0315$ and presentation attenuation of $0.0191$ over raw scaled to matched calibration strength. A scaled raw scorer changes every clean margin and edit response by one common factor. APLOT uniform's positive residuals show smaller held-out edit responses than that proportional adjustment predicts at the fitted clean-margin strength. The two banks establish this additional attenuation separately for broad perturbations and reversible presentation changes (Appendices~\ref{app:scalar} and~\ref{app:aplot-formats}).

The comparison also clarifies the roles of the two controls. Distribution matching isolates where supervision is placed during learning; calibrated scaling matches the magnitude produced after learning. APLOT uniform targets retain less clean-margin magnitude than intact targets, whose retention is $0.8940$, while supplying the additional edit attenuation above. The joint profile makes both properties visible at comparable preference accuracy.

A separate BT presentation-edit study examines how attenuation varies across formatting styles. Using the intact and raw references from the independent-assignment study, it measures mean attenuation $0.0386$ across nine edits. The six emphasis, list, separator, and HTML transformations average $0.0520$, extending the response spectrum beyond the three-operation presentation bank (Appendix~\ref{app:edits}).

\section{Discussion and Conclusion}
We introduced assignment geometry to study how target placement and reward objectives jointly shape learned reward signals. Matched supervision budgets and controlled correspondence make the interaction experimentally accessible. Across five reward objectives, intact correspondence consistently retains larger clean preference margins, while the objective changes the attenuation ordering of target placements. These reward-model comparisons show how a common supervision budget produces different rewards at comparable preference accuracy. Independent reassignments and a related source construction establish the retention direction across changes to both the receiving mapping and the source values.

These findings give reward design a concrete sequence: compare target placements under the intended objective, examine their margin and edit-response profiles within an accuracy budget, and measure extra attenuation against a calibrated reward scale. APLOT uniform targets demonstrate the last step, delivering additional attenuation over independently calibrated raw scaling on both edit banks. Target placement, objective choice, and reward scaling together determine the combinations available to a practitioner.

A next research direction is to learn target assignments from validation profiles and evaluate the resulting reward models in downstream policy optimization. This would connect controlled changes in reward margins and edit responses to generated behavior. Assignment geometry provides the interventions for that study, making where soft targets place their mass an explicit part of reward design.

\bibliography{references}
\bibliographystyle{preprint_references}
\clearpage
\appendix
\setlength{\LTcapwidth}{\linewidth}
\section{Training and Evaluation Details}
\label{app:protocol}
\subsection{Data and populations}
The preference dataset contains 148,211 pairs: 99,926 training, 7,697 validation, 15,782 test, and 24,806 confirmation pairs before prompt-overlap removal. The data catalog comprises UltraFeedback, UltraFeedback-Binarized, HelpSteer, HelpSteer2, RM-Bench, and RewardBench. The confirmation partition before exclusion contains 11,300 UltraFeedback pairs, 10,842 UltraFeedback-Binarized pairs, 1,351 HelpSteer pairs, and 1,313 HelpSteer2 pairs. The two UltraFeedback variants share a provenance family in family-equal sensitivity analyses.

The five-objective main comparison and reward-model extensions P1/P2 use the same strict confirmation population: 22,561 pairs on 12,554 normalized prompt clusters. Normalization applies Unicode NFKC normalization, lowercase conversion, whitespace collapse, and trimming. We exclude every confirmation row whose normalized prompt occurs in training, validation, or test. The respective overlap counts are 121, 904, and 1,252 rows (82, 861, and 1,193 prompts). These sets overlap: their union removes 2,245 rows and 2,108 prompts from the original 24,806 pairs and 14,662 normalized clusters. All primary summaries cluster normalized prompts and hold the population fixed across paired configurations.

The scalar-control analysis uses a deterministic, prompt-disjoint calibration/evaluation split of the strict reward-model population: 11,026/11,535 pairs on 6,145/6,409 prompts.

The main comparison and P1/P2 apply the same prompt exclusion to both edit banks. Aggregate edits retain 21,360 rows on 1,720 normalized prompt clusters from 24,000 rows on 1,940 clusters; presentation edits retain 10,680 rows on the same 1,720 clusters from 12,000 rows on 1,940 clusters. Edits are applied to each response side and paired with that response's original text. Length operations extend a response or truncate it to approximately 55\%; sentiment and agreement operations append positive/negative or agreement/disagreement framing. The presentation bank applies reversible blockquote, heading, and label transformations to 2,000 base pairs before exclusion. The nine-edit presentation study evaluates 36,000 rows on these base pairs before exclusion, using the six raw/intact reference models shared with the independent-assignment study.

\subsection{Source targets and matched interventions}
The primary target mixes a hard preference with the LCC anchor probability $a_i$:
\begin{equation}
t_i=(1-\alpha)+\alpha a_i,\qquad \alpha=0.10,\qquad a_i\in[0.05,0.95].
\end{equation}
Thus the mixed target lies in $[0.905,0.995]$. All pairs receive unit loss weight. The source constructor uses frozen DeBERTa-v3-base representations of prompt--response pairs, averaging final-layer response-token states before taking the chosen-minus-rejected difference. Training-only truncated SVD reduces these differences to 64 components, scaled by their training standard deviations without centering. The fitted design joins this representation block with length, sentiment, and agreement differences and their source interactions. Mirrored feature vectors and labels train a logistic regression with $C=1$, no intercept, and a 1,000-iteration limit. Five prompt-group folds generate out-of-fold training logits; a fit on all training pairs predicts the other splits. The reducer and feature scaling are fit on training data and shared across the five folds.

Let $q_i$ denote the reduced, scaled representation difference and $w_q^{(-f(i))}$ the representation coefficient block fit outside training fold $f(i)$. The anchor is
\begin{equation}
z_i=q_i^\top w_q^{(-f(i))},\qquad
T=\max\!\left\{\frac{Q_{0.9}(|z_{\rm train}|)}{\operatorname{logit}(.9)},.001\right\},\qquad
a_i=\operatorname{clip}_{[.05,.95]}\!\left(\sigma(z_i/T)\right).
\end{equation}
Here $Q_{0.9}$ denotes the empirical 90th percentile. For held-out pairs, $w_q$ comes from the fit on all training pairs. Surface variables enter the logistic fit; the anchor logit uses the representation block alone. Length counts regex words; sentiment uses lexicon polarity counts divided by square-root word count; agreement combines prompt-response token overlap and agreement indicators. The related source variant uses length as the explicit surface feature with source moderation, retaining the same representation-based construction and mixing coefficient.

Strata are defined by split and source label. Source labels identify subdivisions within datasets and can span datasets: the shared label in HelpSteer and HelpSteer2 places their pairs in the same stratum. Within each stratum, Equation~\ref{eq:geometry} preserves the mean correction mass exactly. The primary training mean mass is $0.040508$. Its intact target standard deviation is $0.021173$, while stratum-uniform targets have overall standard deviation $0.003044$ due to differences between stratum means. The second source has mean mass $0.040503$ and intact standard deviation $0.021011$. Mass matching is exact within each source's factorial.

The partial-reassignment operator chooses the specified fraction of rows within each stratum and cyclically shifts their values. The independent-assignment study (P1) generates three within-stratum random orderings and applies a one-position cyclic shift, crossed with the three optimization seeds. Every operator preserves the target multiset on the reassigned subset. P1 shares one raw and one intact reference per seed across all realizations. The second-source study (P2) trains its own uniform, intact, and reassigned configurations and shares P1's raw references. These six reference models define the assignment and source comparisons separately from the twelve matched main BT models.

\subsection{Reward-model budgets and objective implementations}
The reward backbone is DeBERTa-v3-base with a scalar sequence-classification head, implemented in PyTorch with Hugging Face Transformers. Matched BT training uses an NVIDIA RTX PRO 6000 Blackwell GPU with 96 GB memory. Scoring encodes the prompt and response as separate sequences, with longest-first truncation to a total of 512 tokens including special tokens. Training uses one epoch, learning rate $2\times10^{-5}$, warmup fraction $0.03$, weight decay $0.01$, and BF16. Matched main BT trains raw, uniform, intact, and reassigned configurations at seeds $7$, $17$, and $29$. Every configuration uses local batch 16 with accumulation 2, effective batch 32, and 3,123 updates over 99,926 training pairs. The random seed is set before loading the model and initializing its reward head. Evaluation uses the final-epoch model.

The shared raw/intact references and P1/P2 use the same batch size and accumulation as main BT; the references also match its learning rate, one-epoch schedule, and final-epoch selection. The external-objective comparison uses the same one-epoch schedule and nominal effective batch 32; DARM uses local batch 4 with accumulation 8 for its context comparisons.

\begin{table}[h]
\centering\tabfont\setlength{\tabcolsep}{6pt}
\caption{\textbf{Reward-model coverage across study designs.} The 84 distinct models include 60 in the main comparison and 24 in the assignment/source studies.}
\label{tab:checkpoint-accounting}
\begin{tabular}{p{0.30\linewidth}p{0.49\linewidth}r}
\toprule
Study & Model design & Count \\
\midrule
Main matched BT & Four configurations, three seeds & 12 \\
Main external objectives & Four objectives, four configurations, three seeds & 48 \\
Main subtotal & Matched BT and four external objectives & 60 \\
\midrule
Shared BT references & Raw and intact, three seeds & 6 \\
P1 independent assignments & Three reassignment realizations, three seeds & 9 \\
P2 related source variant & Three configurations, three seeds & 9 \\
\midrule
All listed studies & Distinct reward models & 84 \\
\bottomrule
\end{tabular}
\end{table}

\paragraph{BT.} The scalar chosen-minus-rejected reward margin enters the soft binary cross-entropy in Equation~\ref{eq:bt}.
\paragraph{NormBT.} The pairwise loss is multiplied by a detached ratio of the exponential-moving-average representation distance to the current pair's distance. Distances use first-token final-layer representations; EMA decay is $0.9$, with denominator offset $10^{-6}$.
\paragraph{BSR.} The loss adds $0.001 B$ times the square of the batch-mean reward over both response sides, where $B$ is the configured local batch size. At $B=16$ this coefficient is $0.016$.
\paragraph{APLOT.} The reward cost follows the authors' released implementation, $C^{\rm reward}_{ij}=1-\sigma(|r(x_i,y_i^+)-r(x_j,y_j^-)|)$ \citep{aplotcode}. We combine this term with weight $0.9$ and cosine similarity of DeBERTa first-token final-layer representations with weight $0.1$. Uniform transport marginals, entropic regularization $0.1$, and 50 Sinkhorn iterations produce a row-wise adaptive cost that is subtracted from the pair margin before soft cross-entropy. The transport margin is detached during reward optimization.
\paragraph{DARM.} The preference loss is augmented by a context-discrimination loss, weighted $0.05$, on both response sides. Each response is paired with its true prompt and three masked alternatives. Prompt words are divided into eight contiguous spans to construct the alternatives; the contrastive temperature is one.

These common-backbone implementations apply the respective objective mechanisms to the shared target inputs. The normalized-profile comparison excludes the DIR adaptation because its raw margins are near zero at all three seeds. This adaptation uses a categorical CLUB estimator for response length.

\section{Assignment Construction and Target-Logit Geometry}
\label{app:proof}
\subsection{Mean-preserving assignment}
Equation~\ref{eq:geometry} combines dispersion control with within-stratum reassignment. The intermediate mass vector $\widetilde m_i=\bar m_{g(i)}+\lambda(m_i-\bar m_{g(i)})$ contracts deviations from the stratum mean, and $\Pi_p$ reassigns its entries within strata. The shared stratum mean makes this ordering equivalent to the construction in the main text. Fixed response pairs produce model margins independently of the target transformation.

\subsection{Target-logit response}
For a single pair with target $t=1-m$, differentiation of the soft BT loss gives
\begin{equation}
\frac{\partial \ell_{\rm BT}}{\partial d}=\sigma(d)-(1-m).
\end{equation}
For $0<m<1/2$, its unique finite minimizer is $f(m)=\log[(1-m)/m]$. The derivatives are
\begin{equation}
f'(m)=-\frac{1}{m(1-m)},\qquad
f''(m)=\frac{1-2m}{m^2(1-m)^2}>0.
\end{equation}
Applying Jensen's inequality within a stratum gives Equation~\ref{eq:jensen}. Strict convexity gives equality only at constant correction mass. For any fixed set of baseline logits $b_i$, the mean difference $n_g^{-1}\sum_i[b_i-f(m_i)]$ is therefore greatest at the stratum-uniform allocation with the same mean mass.

The interpolation $m_i(\lambda)=\bar m_g+\lambda(m_i-\bar m_g)$ remains in the feasible interval and preserves the stratum mean. Let $F(\lambda)=n_g^{-1}\sum_i f(m_i(\lambda))$. Then $F'(0)=0$ and $F''(\lambda)\geq0$, so $F$ is nondecreasing on $[0,1]$. This describes the average target logit encoded by dispersion. Within-stratum permutations preserve $F$ at fixed $\lambda$ while changing the correspondence of target values to pairs. These two operations therefore provide distinct interventions for the shared-parameter experiments.

\FloatBarrier
\section{Paired Statistical Analysis}
\label{app:inference}
\paragraph{Estimands and sampling.}
The unit of evaluation aggregation is a normalized prompt cluster. We average observations within a prompt and then average prompts equally; each training seed contributes equally. A positive clean margin counts as a correct preference prediction, with zero margins counted as incorrect. The raw scale $s_0$ is the row-average absolute clean margin on strict confirmation for all five reward objectives and P1/P2. Retention adds the prompt-average paired magnitude change to one, as in Equation~\ref{eq:retention}.

The main comparison and P1/P2 use 50,000 crossed bootstrap draws. Each draw samples prompt clusters with replacement and shares their multiplicities across matched configurations, seeds, and assignment realizations. Clean and edit banks share these weights wherever prompts coincide, and each bank retains its own prompt-average denominator. Seed weights are sampled independently of prompt weights. P1 also samples independent weights for its three assignment realizations, preserving the complete realization-by-seed crossing and the shared raw/intact references. For each objective and seed $k$, the raw scale is recomputed in every profile draw:
\begin{equation}
s_{0,k}^{*}=\frac{\sum_P w_P^{*}\sum_{j\in P}|d_{0,k,j}|}{\sum_P w_P^{*}n_P},
\label{eq:bootstrap-scale}
\end{equation}
where the sums range over clean prompts and $n_P$ is their pair count. Profiles and their direct contrasts are calculated before averaging seeds and, for P1, assignments with their sampled weights. Raw retention remains exactly one in every draw.

Seed resampling uses the empirical distribution of the three observed training realizations. The resulting intervals combine prompt variation with this empirical seed variation. Appendix~\ref{app:sampling-sensitivity} distinguishes this estimand from inference conditional on the three scorers and reports leave-one-seed sensitivity for APLOT Uniform. The number of independent training realizations remains three.

\paragraph{Interval labels and comparison families.}
The three contrasts in Equation~\ref{eq:contrasts} are denoted H1, H2, and H3, respectively. The five reward objectives share a 15-contrast H family. The two P1 and three P2 contrasts retain a conservative correction factor $K=7$. Assignment-specific P1 intervals resample prompts and seeds while fixing the displayed assignment, with $K=21$; the pooled P1 comparison also resamples assignments. A one-sided familywise 95\% lower bound uses bootstrap quantile $.05/K$; a two-sided familywise 95\% interval uses quantiles $.025/K$ and $1-.025/K$. A contrast passes the stated effect criterion when its one-sided familywise lower bound exceeds $0.01$. Tables~\ref{tab:objective-intervals} and~\ref{tab:replication-intervals} display the full two-sided intervals for the gains summarized in main Tables~\ref{tab:objectives} and~\ref{tab:replication}. Marginal intervals use the unadjusted $.025$ and $.975$ quantiles and are explicitly labeled.

Profile and utility families are defined separately for each endpoint and group: three candidates per main objective and for P2, and two for P1. The expanded-edit mean uses 20,000 prompt-and-seed draws. Individual edit estimates provide a descriptive breakdown; their complete intervals are included with the accompanying numerical data.

\paragraph{Calibrated scalar comparisons.}
BT and APLOT use 50,000 crossed draws and a 15-comparison family per objective. Calibration and evaluation prompts are sampled independently, with shared weights across configurations, seeds, and overlapping clean/edit observations within each partition. Seed weights are sampled independently of prompt weights. Each draw refits the coefficient in Equation~\ref{eq:scalar-fit}, recomputes the raw row-average scale over both partitions using Equation~\ref{eq:bootstrap-scale}, and averages seed-specific residuals.

\paragraph{Crossed objective comparisons.}
The interaction analysis combines matched BT with NormBT, BSR, APLOT, and DARM on the common strict clean and edit populations. The same 50,000 crossed draws share prompt and seed weights across every objective and configuration. An interaction is the difference of the same geometry contrast between two objectives. The ten objective pairs and three contrasts define a 30-comparison family. Table~\ref{tab:crossed} reports every interaction with its two-sided familywise 95\% interval.

\paragraph{Accuracy equivalence.}
Utility equivalence uses the symmetric interval $[-0.01,0.01]$. A candidate is equivalent to its raw control when both two-sided familywise interval endpoints lie inside this interval. The comparison establishes the accuracy budget within which the continuous attenuation-retention profiles are interpreted.

\section{Matched BT Profile}
\label{app:matched-bt}
The main BT comparison uses matched seed initializations and a common training schedule for raw, uniform, intact, and reassigned targets. Table~\ref{tab:bt-profile-main} reports the mean profiles; Tables~\ref{tab:objective-intervals} and~\ref{tab:utility-all} give the paired contrasts and accuracy-equivalence results.

\FloatBarrier
\section{Objective-Specific Retention and Interactions}
\label{app:interactions}
\begin{table}[h]
\centering\tabfont\setlength{\tabcolsep}{3pt}
\caption{\textbf{Geometry contrasts across reward objectives.} Estimates and two-sided familywise 95\% intervals follow Appendix~\ref{app:inference}. Positive $\Delta_{p,R}$ favors intact correspondence.}
\label{tab:objective-intervals}
\renewcommand{\arraystretch}{1.18}
\begin{tabular}{lccc}
\toprule
Objective & Dispersion $\Delta_\lambda$ & Reassignment $\Delta_{p,A}$ & Retention $\Delta_{p,R}$ \\
\midrule
BT & \shortstack{$+0.0304$ \\ $[+0.0123, +0.0542]$} & \shortstack{$+0.0646$ \\ $[+0.0298, +0.0941]$} & \shortstack{$+0.0779$ \\ $[+0.0666, +0.0929]$} \\
NormBT & \shortstack{$-0.0334$ \\ $[-0.0563, -0.0075]$} & \shortstack{$-0.0144$ \\ $[-0.1013, +0.0356]$} & \shortstack{$+0.0639$ \\ $[+0.0588, +0.0690]$} \\
BSR & \shortstack{$-0.0139$ \\ $[-0.0947, +0.0386]$} & \shortstack{$+0.0203$ \\ $[-0.0274, +0.0561]$} & \shortstack{$+0.0946$ \\ $[+0.0711, +0.1296]$} \\
APLOT & \shortstack{$+0.0670$ \\ $[+0.0394, +0.0871]$} & \shortstack{$+0.0514$ \\ $[+0.0215, +0.0910]$} & \shortstack{$+0.0674$ \\ $[+0.0520, +0.0834]$} \\
DARM & \shortstack{$+0.0176$ \\ $[-0.0215, +0.0437]$} & \shortstack{$+0.0459$ \\ $[-0.0343, +0.1281]$} & \shortstack{$+0.0783$ \\ $[+0.0195, +0.1109]$} \\
\bottomrule
\end{tabular}
\end{table}
Intact correspondence retains larger clean margins across all five objectives, while dispersion changes attenuation in opposite directions under NormBT and under BT or APLOT. The direct interactions in Table~\ref{tab:crossed} quantify these objective-dependent differences.
\begingroup
\tabfont\setlength{\tabcolsep}{4pt}\renewcommand{\arraystretch}{1.08}
\setlength{\LTcapwidth}{\textwidth}
\begin{longtable}{llrr}
\caption{\textbf{All 30 direct objective interactions.} Each estimate is the first objective's contrast minus the second's. H1--H3 follow Equation~\ref{eq:contrasts}; two-sided familywise 95\% intervals follow Appendix~\ref{app:inference}.}\label{tab:crossed} \\
\toprule
Objective comparison & Contrast & Estimate & Familywise 95\% interval \\
\midrule
\endfirsthead
\multicolumn{4}{l}{\textit{Continued from previous page}} \\
\toprule
Objective comparison & Contrast & Estimate & Familywise 95\% interval \\
\midrule
\endhead
\midrule\multicolumn{4}{r}{\textit{Continued on next page}} \\
\endfoot
\bottomrule
\endlastfoot
BT $-$ NormBT & H1 & $+0.0638$ & $[+0.0216, +0.1096]$ \\
BT $-$ NormBT & H2 & $+0.0791$ & $[-0.0053, +0.1752]$ \\
BT $-$ NormBT & H3 & $+0.0139$ & $[+0.0002, +0.0329]$ \\
\addlinespace[2pt]
BT $-$ BSR & H1 & $+0.0443$ & $[-0.0146, +0.1483]$ \\
BT $-$ BSR & H2 & $+0.0444$ & $[-0.0246, +0.1017]$ \\
BT $-$ BSR & H3 & $-0.0167$ & $[-0.0617, +0.0114]$ \\
\addlinespace[2pt]
BT $-$ APLOT & H1 & $-0.0366$ & $[-0.0663, -0.0127]$ \\
BT $-$ APLOT & H2 & $+0.0133$ & $[-0.0225, +0.0711]$ \\
BT $-$ APLOT & H3 & $+0.0105$ & $[-0.0155, +0.0396]$ \\
\addlinespace[2pt]
BT $-$ DARM & H1 & $+0.0128$ & $[-0.0123, +0.0395]$ \\
BT $-$ DARM & H2 & $+0.0187$ & $[-0.0589, +0.1276]$ \\
BT $-$ DARM & H3 & $-0.0004$ & $[-0.0348, +0.0509]$ \\
\addlinespace[2pt]
NormBT $-$ BSR & H1 & $-0.0195$ & $[-0.0782, +0.0439]$ \\
NormBT $-$ BSR & H2 & $-0.0347$ & $[-0.0796, -0.0026]$ \\
NormBT $-$ BSR & H3 & $-0.0307$ & $[-0.0665, -0.0053]$ \\
\addlinespace[2pt]
NormBT $-$ APLOT & H1 & $-0.1004$ & $[-0.1419, -0.0728]$ \\
NormBT $-$ APLOT & H2 & $-0.0658$ & $[-0.1915, +0.0023]$ \\
NormBT $-$ APLOT & H3 & $-0.0035$ & $[-0.0203, +0.0127]$ \\
\addlinespace[2pt]
NormBT $-$ DARM & H1 & $-0.0511$ & $[-0.0993, +0.0122]$ \\
NormBT $-$ DARM & H2 & $-0.0604$ & $[-0.2295, +0.0585]$ \\
NormBT $-$ DARM & H3 & $-0.0144$ & $[-0.0503, +0.0469]$ \\
\addlinespace[2pt]
BSR $-$ APLOT & H1 & $-0.0809$ & $[-0.1798, -0.0024]$ \\
BSR $-$ APLOT & H2 & $-0.0311$ & $[-0.1168, +0.0160]$ \\
BSR $-$ APLOT & H3 & $+0.0272$ & $[+0.0025, +0.0520]$ \\
\addlinespace[2pt]
BSR $-$ DARM & H1 & $-0.0315$ & $[-0.1372, +0.0372]$ \\
BSR $-$ DARM & H2 & $-0.0257$ & $[-0.1539, +0.0690]$ \\
BSR $-$ DARM & H3 & $+0.0163$ & $[-0.0357, +0.1091]$ \\
\addlinespace[2pt]
APLOT $-$ DARM & H1 & $+0.0493$ & $[+0.0042, +0.1001]$ \\
APLOT $-$ DARM & H2 & $+0.0054$ & $[-0.0425, +0.0617]$ \\
APLOT $-$ DARM & H3 & $-0.0109$ & $[-0.0575, +0.0627]$ \\
\end{longtable}
\endgroup

\FloatBarrier
\section{Independent Assignments and a Related Source Variant}
\label{app:assignments}
To examine retention across assignment maps and source constructions, P1 crosses three independent within-stratum reassignments with three optimization seeds, sharing raw and intact references across realizations. P2 uses a related source constructor with the same stratum-matched interventions. Table~\ref{tab:replication-intervals} reports their retention and attenuation contrasts on the strict evaluation populations.
\begin{table}[ht]
\centering\tabfont\setlength{\tabcolsep}{7pt}\renewcommand{\arraystretch}{1.08}
\caption{\textbf{Assignment and source contrasts.} H1--H3 follow Equation~\ref{eq:contrasts}. Each study uses its own paired references on the strict population; two-sided familywise 95\% intervals follow Appendix~\ref{app:inference}.}
\label{tab:replication-intervals}
\begin{tabular}{llrr}
\toprule
Study & Contrast & Estimate & Familywise 95\% interval \\
\midrule
Independent assignments & H2 & $+0.0355$ & $[-0.0325, +0.0930]$ \\
Independent assignments & H3 & $+0.0761$ & $[+0.0616, +0.0911]$ \\
\addlinespace
Related source variant & H1 & $+0.0306$ & $[-0.0116, +0.0850]$ \\
Related source variant & H2 & $+0.0386$ & $[+0.0115, +0.0614]$ \\
Related source variant & H3 & $+0.0885$ & $[+0.0708, +0.1194]$ \\
\addlinespace
\bottomrule
\end{tabular}
\end{table}

The retention gain appears in both studies and in each P1 assignment realization, detailed in Table~\ref{tab:assignment-replication}.
\begin{table}[ht]
\centering\tabfont\setlength{\tabcolsep}{4pt}\renewcommand{\arraystretch}{1.08}
\caption{\textbf{P1 retention effects by independent assignment realization.} H3 is intact-minus-reassigned clean-margin retention on strict confirmation. Assignment-specific two-sided familywise 95\% intervals follow Appendix~\ref{app:inference}.}
\label{tab:assignment-replication}
\begin{tabular}{lrrr}
\toprule
Assignment & H3 & Familywise 95\% interval & $K$ \\
\midrule
101 & $+0.0756$ & $[+0.0556, +0.0945]$ & 21 \\
202 & $+0.0813$ & $[+0.0674, +0.0934]$ & 21 \\
303 & $+0.0714$ & $[+0.0649, +0.0780]$ & 21 \\
\bottomrule
\end{tabular}
\end{table}

\FloatBarrier
\section{Utility Equivalence}
\label{app:utility}
All fifteen main configurations satisfy the symmetric one-percentage-point accuracy-equivalence criterion relative to their objective's raw control, as reported in Table~\ref{tab:utility-all}. Table~\ref{tab:extension-utility} gives the corresponding results for P1 and P2; all five comparisons satisfy the same criterion. P1's reassigned result averages assignment realizations and seeds. Each study uses the paired references in Appendix~\ref{app:protocol} and comparison families in Appendix~\ref{app:inference}.
\begin{table}[ht]
\centering\tabfont\setlength{\tabcolsep}{4pt}\renewcommand{\arraystretch}{1.08}
\caption{\textbf{Accuracy relative to each objective's own raw control.} Differences and two-sided familywise 95\% intervals are percentage points. All intervals lie inside $[-1,1]$. Comparison families follow Appendix~\ref{app:inference}.}
\label{tab:utility-all}
\begin{tabular}{llrr}
\toprule
Objective & Target & Difference (pp) & Familywise interval (pp) \\
\midrule
BT & Uniform & $-0.091$ & $[-0.663, +0.490]$ \\
BT & Intact & $+0.003$ & $[-0.556, +0.516]$ \\
BT & Reassigned & $+0.067$ & $[-0.427, +0.597]$ \\
\addlinespace
NormBT & Uniform & $-0.063$ & $[-0.454, +0.335]$ \\
NormBT & Intact & $+0.117$ & $[-0.353, +0.603]$ \\
NormBT & Reassigned & $-0.104$ & $[-0.659, +0.410]$ \\
\addlinespace
BSR & Uniform & $+0.189$ & $[-0.241, +0.596]$ \\
BSR & Intact & $+0.146$ & $[-0.305, +0.599]$ \\
BSR & Reassigned & $+0.049$ & $[-0.389, +0.501]$ \\
\addlinespace
APLOT & Uniform & $-0.225$ & $[-0.639, +0.187]$ \\
APLOT & Intact & $-0.050$ & $[-0.486, +0.393]$ \\
APLOT & Reassigned & $-0.022$ & $[-0.458, +0.412]$ \\
\addlinespace
DARM & Uniform & $+0.047$ & $[-0.459, +0.544]$ \\
DARM & Intact & $+0.127$ & $[-0.576, +0.976]$ \\
DARM & Reassigned & $-0.205$ & $[-0.806, +0.335]$ \\
\bottomrule
\end{tabular}
\end{table}

\begin{table}[ht]
\centering\tabfont\setlength{\tabcolsep}{4pt}\renewcommand{\arraystretch}{1.08}
\caption{\textbf{Accuracy equivalence across assignments and source constructions.} Differences and two-sided familywise 95\% intervals are percentage points relative to each study's raw control. P1 reassigned averages three assignment realizations and three seeds. All intervals lie inside $[-1,1]$.}
\label{tab:extension-utility}
\begin{tabular}{llrr}
\toprule
Study & Target & Difference (pp) & Familywise interval (pp) \\
\midrule
Independent assignments & Intact & $+0.058$ & $[-0.486, +0.583]$ \\
Independent assignments & Reassigned & $-0.009$ & $[-0.583, +0.604]$ \\
\addlinespace
Related source variant & Uniform & $+0.226$ & $[-0.283, +0.800]$ \\
Related source variant & Intact & $+0.148$ & $[-0.515, +0.743]$ \\
Related source variant & Reassigned & $-0.123$ & $[-0.820, +0.573]$ \\
\bottomrule
\end{tabular}
\end{table}

\FloatBarrier
\section{Absolute Raw Metrics and Seed Sensitivity}
\label{app:seed-sensitivity}
Table~\ref{tab:raw-metrics} provides the absolute accuracy and margin scales underlying the normalized profiles on strict confirmation.
\begingroup
\tabfont\setlength{\tabcolsep}{4pt}\renewcommand{\arraystretch}{1.08}
\setlength{\LTcapwidth}{\textwidth}
\begin{longtable}{lrrrr}
\caption{\textbf{Absolute raw accuracy and margin scale by training seed.} Pair accuracy weights clean pairs equally; prompt accuracy averages pairs within each prompt and then weights prompts equally. Both are percentages. Margin scale is the raw pair-average absolute clean margin on strict confirmation. P1 and P2 share the same raw reference model at each seed.}\label{tab:raw-metrics} \\
\toprule
Group & Seed & Pair accuracy (\%) & Prompt accuracy (\%) & Margin scale \\
\midrule
\endfirsthead
\multicolumn{5}{l}{\textit{Continued from previous page}} \\
\toprule
Group & Seed & Pair accuracy (\%) & Prompt accuracy (\%) & Margin scale \\
\midrule
\endhead
\midrule\multicolumn{5}{r}{\textit{Continued on next page}} \\
\endfoot
\bottomrule
\endlastfoot
BT & 7 & $74.32$ & $73.36$ & $1.4462$ \\
BT & 17 & $74.88$ & $73.98$ & $1.5677$ \\
BT & 29 & $74.55$ & $73.75$ & $1.4561$ \\
\addlinespace
NormBT & 7 & $74.53$ & $73.61$ & $1.3821$ \\
NormBT & 17 & $74.90$ & $74.01$ & $1.5237$ \\
NormBT & 29 & $74.49$ & $73.60$ & $1.4313$ \\
\addlinespace
BSR & 7 & $74.56$ & $73.74$ & $1.3924$ \\
BSR & 17 & $74.72$ & $73.77$ & $1.4719$ \\
BSR & 29 & $74.51$ & $73.61$ & $1.4107$ \\
\addlinespace
APLOT & 7 & $73.78$ & $72.66$ & $1.2249$ \\
APLOT & 17 & $73.79$ & $72.61$ & $1.2897$ \\
APLOT & 29 & $73.62$ & $72.53$ & $1.2909$ \\
\addlinespace
DARM & 7 & $73.42$ & $72.35$ & $1.2925$ \\
DARM & 17 & $73.40$ & $72.24$ & $1.2643$ \\
DARM & 29 & $73.90$ & $72.73$ & $1.2791$ \\
\addlinespace
P1/P2 & 7 & $74.33$ & $73.39$ & $1.4166$ \\
P1/P2 & 17 & $74.74$ & $73.79$ & $1.5385$ \\
P1/P2 & 29 & $74.90$ & $74.02$ & $1.4734$ \\
\end{longtable}
\endgroup

Table~\ref{tab:seed-points} separates all three contrasts by training seed. The retention gain is positive in all fifteen observed objective-seed combinations; aggregate inference appears in Table~\ref{tab:objective-intervals}.
\begin{table}[ht]
\centering\tabfont\setlength{\tabcolsep}{12pt}\renewcommand{\arraystretch}{1.08}
\caption{\textbf{Geometry contrasts at each training seed.} All five objectives use the same strict evaluation population. H1, H2, and H3 follow Equation~\ref{eq:contrasts}. Aggregate estimates and crossed-bootstrap intervals appear in Table~\ref{tab:objective-intervals}.}
\label{tab:seed-points}
\begin{tabular}{llrrr}
\toprule
Objective & Contrast & Seed 7 & Seed 17 & Seed 29 \\
\midrule
BT & H1 & $+0.0155$ & $+0.0251$ & $+0.0506$ \\
BT & H2 & $+0.0903$ & $+0.0329$ & $+0.0707$ \\
BT & H3 & $+0.0690$ & $+0.0901$ & $+0.0746$ \\
\addlinespace
NormBT & H1 & $-0.0119$ & $-0.0365$ & $-0.0518$ \\
NormBT & H2 & $+0.0219$ & $+0.0314$ & $-0.0967$ \\
NormBT & H3 & $+0.0646$ & $+0.0625$ & $+0.0648$ \\
\addlinespace
BSR & H1 & $+0.0131$ & $+0.0353$ & $-0.0901$ \\
BSR & H2 & $+0.0311$ & $+0.0528$ & $-0.0231$ \\
BSR & H3 & $+0.1261$ & $+0.0832$ & $+0.0746$ \\
\addlinespace
APLOT & H1 & $+0.0758$ & $+0.0427$ & $+0.0824$ \\
APLOT & H2 & $+0.0254$ & $+0.0422$ & $+0.0865$ \\
APLOT & H3 & $+0.0800$ & $+0.0556$ & $+0.0667$ \\
\addlinespace
DARM & H1 & $-0.0183$ & $+0.0311$ & $+0.0402$ \\
DARM & H2 & $-0.0308$ & $+0.0458$ & $+0.1228$ \\
DARM & H3 & $+0.0230$ & $+0.1073$ & $+0.1046$ \\
\addlinespace
\bottomrule
\end{tabular}
\end{table}

\FloatBarrier
\section{Calibrated Scalar Controls}
\label{app:scalar}
\paragraph{Analytic scaling profile.}
For a positive scalar $c$, let $r_c=cr_0$. Then $d_c=cd_0$ and $b_c(e)=cb_0(e)$, preserving the sign of every clean margin and hence pair accuracy. With the prompt weighting and raw row-average scale used in Equations~\ref{eq:retention} and~\ref{eq:attenuation}, define
\begin{equation}
\kappa=\frac{\E_P|d_0|}{s_0},\qquad
\eta_{\mathcal E}=\frac{\E_{P,e\in\mathcal E}|b_0(e)|}{s_0}.
\end{equation}
The resulting trajectory is
\begin{equation}
\R(c)=1+(c-1)\kappa,\qquad
\A_{\mathcal E}(c)=(1-c)\eta_{\mathcal E},\qquad
\Delta\U(c)=0.
\label{eq:scalar-profile}
\end{equation}
These identities apply within each seed before averaging. The factor $\kappa$ accounts for the distinct prompt-average and row-average weights.

\paragraph{Independent calibration and inference.}
For candidate $\theta$ and base $b$, the fitted coefficient is
\begin{equation}
\widehat c_{b,\theta}
=\frac{\E_{P\in\mathcal C}|d_\theta|}{\E_{P\in\mathcal C}|d_b|},
\label{eq:scalar-fit}
\end{equation}
where $\mathcal C$ is the independent clean calibration partition. Coefficients are fit separately for every seed using clean margins alone. Clean and edit evaluation use the disjoint evaluation prompts specified in Appendix~\ref{app:protocol}.

Extra edit attenuation is $\E_P[\widehat c_{b,\theta}|b_b(e)|-|b_\theta(e)|]/s_0$. Evaluation retention mismatch is $\E_P[|d_\theta|-\widehat c_{b,\theta}|d_b|]/s_0$, measuring how the calibration match transfers to held-out pairs. Raw is scaled to each candidate, and intact is additionally scaled to uniform and reassigned. The five comparisons each yield two edit endpoints and retention mismatch; inference follows Appendix~\ref{app:inference}.
\begin{table}[ht]
\centering\tabfont\setlength{\tabcolsep}{7pt}\renewcommand{\arraystretch}{1.08}
\caption{\textbf{Attenuation beyond calibrated scalar controls.} Each candidate is compared with raw scaled to match the candidate's clean absolute margin on calibration prompts. Positive residuals denote extra attenuation. Two-sided familywise 95\% intervals follow Appendix~\ref{app:inference}.}
\label{tab:scalar-controls}
\begin{tabular}{llrr}
\toprule
Target & Response & Extra attenuation & Familywise 95\% interval \\
\midrule
\multicolumn{4}{l}{\textit{BT}} \\
Uniform & Aggregate & $-0.01432$ & $[-0.03625, +0.00419]$ \\
Uniform & Presentation & $+0.00426$ & $[-0.02125, +0.02728]$ \\
Intact & Aggregate & $-0.01272$ & $[-0.03184, +0.01275]$ \\
Intact & Presentation & $+0.01511$ & $[-0.00272, +0.03957]$ \\
Reassigned & Aggregate & $+0.01585$ & $[-0.00069, +0.04109]$ \\
Reassigned & Presentation & $+0.01289$ & $[-0.00754, +0.04105]$ \\
\addlinespace
\multicolumn{4}{l}{\textit{APLOT}} \\
Uniform & Aggregate & $+0.03153$ & $[+0.00637, +0.06602]$ \\
Uniform & Presentation & $+0.01914$ & $[+0.00592, +0.03138]$ \\
Intact & Aggregate & $-0.01920$ & $[-0.05949, +0.01999]$ \\
Intact & Presentation & $+0.00468$ & $[-0.03030, +0.02573]$ \\
Reassigned & Aggregate & $-0.00034$ & $[-0.04060, +0.03402]$ \\
Reassigned & Presentation & $+0.01584$ & $[-0.00177, +0.03495]$ \\
\addlinespace
\bottomrule
\end{tabular}
\end{table}

The paired comparisons in Table~\ref{tab:scalar-controls} identify APLOT uniform as a combination with extra attenuation on both edit banks. Its held-out retention mismatch is $-0.0044$, with two-sided familywise 95\% interval $[-0.0194,0.0113]$.

\clearpage
\section{Presentation-Edit Spectrum}
\label{app:edits}
The presentation transformations retain an exact inverse at the string level. Evaluation scores the original response against its transformed counterpart with the prompt held fixed. The three-operation bank uses blockquote prefixes, a section heading, and a response label. The expanded bank additionally uses bold and italic wrappers, ordered and unordered lists, separators, and an HTML section container. Prompt clustering retains all edited versions of each prompt together.

Table~\ref{tab:edits} reports all nine edits using the six raw/intact references shared with P1 and the full edit population. The equal-edit mean attenuation is $0.0386$, with marginal 95\% interval $[0.0075,0.0801]$ under the inference in Appendix~\ref{app:inference}. Mean attenuation is $0.0119$ for the three-operation presentation bank and $0.0520$ for the six other presentation transformations. Individual point estimates describe how the response varies across presentation styles.
\begin{table}[ht]
\centering\tabfont\setlength{\tabcolsep}{12pt}\renewcommand{\arraystretch}{1.08}
\caption{\textbf{Attenuation across nine presentation edits.} Intact relative to raw using the references shared with P1 at three seeds, on the full edit population. Entries are descriptive point estimates; uncertainty for the equal-edit mean is reported in the text.}
\label{tab:edits}
\begin{tabular}{lr}
\toprule
Edit & Attenuation \\
\midrule
Blockquote & $0.0101$ \\
Heading & $0.0240$ \\
HTML section & $0.1258$ \\
Response label & $0.0016$ \\
Bold wrapper & $0.0276$ \\
Italic wrapper & $0.0477$ \\
Ordered list & $0.0386$ \\
Separator & $0.0388$ \\
Unordered list & $0.0336$ \\
\bottomrule
\end{tabular}
\end{table}

\clearpage
\section{APLOT Format Responses and Scaling Detail}
\label{app:aplot-formats}
\subsection{Nine-format comparison on strict prompts}
The APLOT comparison applies the nine transformations in Appendix~\ref{app:edits} to raw, uniform, intact, and reassigned scorers at all three training seeds. The prompt exclusions used in the main comparison retain 32,040 observations on 1,720 prompts. The scalar comparison uses 16,416 observations on the 884 edit prompts in the evaluation partition. Each coefficient is fit on the independent clean calibration partition using Equation~\ref{eq:scalar-fit}; edit responses do not enter the fit.

Table~\ref{tab:aplot-extended} distinguishes raw-relative attenuation on the strict edit population from extra attenuation over calibrated scaling on held-out prompts. Uniform targets yield mean attenuation $0.0354$ relative to raw and mean extra attenuation $0.0111$ relative to scaling. Their two-sided familywise 95\% intervals are $[0.0257,0.0438]$ and $[0.0028,0.0209]$, respectively. The corresponding Intact and Reassigned mean scaling residuals are $0.0054$ and $0.0072$, with intervals $[-0.0097,0.0241]$ and $[-0.0066,0.0240]$. The per-edit entries show how this average response is distributed across formatting operations.

Inference uses 50,000 crossed prompt-and-seed draws. Prompt weights are shared across configurations and seeds, with separate sampling of calibration and evaluation prompts. Each draw refits the clean coefficient and recomputes the strict clean row-average scale. For each reference type, the three configurations and ten summaries, comprising nine edits and their mean, form a 30-comparison family. Individual edit estimates are descriptive; the accompanying numerical data include every interval and seed-specific value.
\begin{table}[h]
\centering\tabfont
\setlength{\tabcolsep}{3pt}\renewcommand{\arraystretch}{1.14}
\caption{\textbf{APLOT responses across nine formatting operations.} All three target assignments are evaluated against their paired raw and calibrated-scaling references.}
\label{tab:aplot-extended}
\begin{tabular}{lrrrrrr}
\toprule
& \multicolumn{3}{c}{Raw-relative attenuation} & \multicolumn{3}{c}{Extra attenuation after scaling} \\
\cmidrule(lr){2-4}\cmidrule(lr){5-7}
Edit & Uniform & Intact & Reassigned & Uniform & Intact & Reassigned \\
\midrule
Blockquote & $0.0165$ & $0.0221$ & $0.0253$ & $-0.0015$ & $0.0095$ & $0.0022$ \\
Heading & $0.0721$ & $0.0363$ & $0.0760$ & $0.0422$ & $0.0133$ & $0.0382$ \\
HTML section & $0.0542$ & $0.0812$ & $0.0949$ & $0.0027$ & $0.0443$ & $0.0314$ \\
Response label & $0.0445$ & $0.0126$ & $0.0420$ & $0.0167$ & $-0.0087$ & $0.0071$ \\
Bold wrapper & $0.0322$ & $0.0024$ & $0.0279$ & $0.0141$ & $-0.0121$ & $0.0059$ \\
Italic wrapper & $0.0492$ & $0.0223$ & $0.0293$ & $0.0270$ & $0.0055$ & $0.0004$ \\
Ordered list & $0.0213$ & $0.0170$ & $0.0203$ & $-0.0022$ & $-0.0002$ & $-0.0098$ \\
Separators & $0.0158$ & $0.0062$ & $0.0102$ & $0.0030$ & $-0.0039$ & $-0.0071$ \\
Unordered list & $0.0124$ & $0.0114$ & $0.0153$ & $-0.0018$ & $0.0010$ & $-0.0035$ \\
\midrule
Nine-edit mean & $0.0354$ & $0.0235$ & $0.0379$ & $0.0111$ & $0.0054$ & $0.0072$ \\
\bottomrule
\end{tabular}
\par\vspace{4pt}
\begin{minipage}{.98\linewidth}\footnotesize Raw-relative estimates use 32,040 strict edit observations on 1,720 prompts. Scaling residuals use the disjoint evaluation subset of 16,416 observations on 884 prompts. Means weight the nine edit types equally. Full intervals and per-seed estimates accompany the numerical data.\end{minipage}
\end{table}

\clearpage
\subsection{Training-seed detail and response units}
Table~\ref{tab:aplot-seed-scaling} expands the APLOT Uniform comparison from Table~\ref{tab:scalar-controls}. Extra attenuation is positive for aggregate edits and for the three-operation presentation bank at each observed training seed. The lower panel reports the reward-score magnitudes before division by $s_0$, making the normalized residuals directly traceable to their paired responses. The clean margin mismatch measures transfer of the independently fitted coefficient to evaluation prompts.
\begin{table}[h]
\centering\tabfont
\setlength{\tabcolsep}{4pt}\renewcommand{\arraystretch}{1.14}
\caption{\textbf{APLOT Uniform scaling comparisons by training seed.} The upper panel reports normalized residuals; the lower panel exposes their underlying response magnitudes.}
\label{tab:aplot-seed-scaling}
\begin{tabular}{lrrrrr}
\toprule
Seed & $\widehat c$ & $s_0$ & Margin mismatch & Aggregate extra & Presentation extra \\
\midrule
7 & $0.8236$ & $1.2249$ & $0.0042$ & $0.0211$ & $0.0098$ \\
17 & $0.8822$ & $1.2897$ & $-0.0060$ & $0.0606$ & $0.0268$ \\
29 & $0.8617$ & $1.2909$ & $-0.0115$ & $0.0129$ & $0.0208$ \\
\midrule
Mean & $0.8558$ & $1.2685$ & $-0.0044$ & $0.0315$ & $0.0191$ \\
\midrule
\multicolumn{6}{l}{\textit{Absolute edit responses in reward-score units}} \\
& \multicolumn{2}{c}{Aggregate} & \multicolumn{2}{c}{Presentation} & \\
Seed & Scaled raw & Uniform & Scaled raw & Uniform &  \\
\midrule
7 & $0.5017$ & $0.4759$ & $0.1671$ & $0.1552$ &  \\
17 & $0.5205$ & $0.4423$ & $0.2036$ & $0.1690$ &  \\
29 & $0.5370$ & $0.5204$ & $0.1990$ & $0.1722$ &  \\
\bottomrule
\end{tabular}
\par\vspace{4pt}
\begin{minipage}{.98\linewidth}\footnotesize Each coefficient is fit on clean calibration prompts. Extra attenuation subtracts the trained response magnitude from its scaled-raw counterpart, then divides by the seed-specific $s_0$. The mean averages these seed-specific residuals. The complete numerical data also report Intact and Reassigned.\end{minipage}
\end{table}

\paragraph{Interpretation by edit type.}
The two edit banks probe different properties of the scoring function. Table~\ref{tab:edit-roles} separates reversible presentation changes from operations that alter response content, tone, or stance. For the aggregate bank, attenuation quantifies sensitivity to those interventions. Presentation transformations preserve the underlying response through a defined string inverse. Both banks use the token-limited scoring inputs specified in Appendix~\ref{app:protocol}. These paired-score measurements use the recorded preferences for clean-pair accuracy; edited responses do not have additional preference or quality labels.
\begin{table}[h]
\centering\tabfont
\setlength{\tabcolsep}{5pt}\renewcommand{\arraystretch}{1.14}
\caption{\textbf{Edit operations and their evaluation roles.} Each edit is scored against the original response under the same prompt.}
\label{tab:edit-roles}
\begin{tabular}{p{.18\linewidth}p{.39\linewidth}p{.34\linewidth}}
\toprule
Group & Operation & Evaluation role \\
\midrule
Presentation & Blockquote, heading, response label; bold and italic wrappers, ordered and unordered lists, separators, HTML section & Sensitivity to reversible formatting. The inverse restores the original response string. \\
\addlinespace
Length & Extend the response or truncate it to approximately 55\% & Sensitivity to response length and the accompanying content change. \\
\addlinespace
Sentiment and agreement & Append positive or negative framing, or agreement or disagreement wording & Sensitivity to changes in tone and stance. \\
\bottomrule
\end{tabular}
\end{table}

\clearpage
\section{Margin Contributions and Calibrated Predictions}
\label{app:margin-diagnostics}
\subsection{Common prediction groups}
The margin analysis uses the 11,535 clean evaluation pairs on 6,409 prompts from the independent calibration split. For each seed, the raw scorer partitions these pairs into raw-correct and raw-incorrect groups using the sign of its preferred-minus-less-preferred margin. Every candidate is evaluated on these same groups. Group membership therefore stays fixed when configurations are compared.

For group $G$, its absolute contribution is the prompt average of $\mathbf1_G|d_\theta|$ divided by the evaluation population's row-average raw magnitude. Its signed contribution replaces $|d_\theta|$ with $d_\theta$. Both averages include all evaluation prompts, assigning zero contribution to observations outside $G$. Consequently, the two group contributions sum to the overall magnitude or signed-margin summary on this population.

Table~\ref{tab:margin-groups} locates most of the Intact--Reassigned magnitude increment on raw-correct pairs across all five objectives. Raw-incorrect pairs also contribute to the increment. The signed entries show positive mean contributions on the raw-correct group and negative mean contributions on the raw-incorrect group, with larger magnitudes under Intact than Reassigned in both groups. This decomposition explains where the retained magnitude lies relative to the recorded preference direction. Predictive losses on the same evaluation population provide the complementary probability assessment below.
\begin{table}[h]
\centering\tabfont
\setlength{\tabcolsep}{4pt}\renewcommand{\arraystretch}{1.14}
\caption{\textbf{Location and direction of the retained-margin increment.} Groups are defined once from each seed\textquotesingle s raw prediction and shared across configurations.}
\label{tab:margin-groups}
\begin{tabular}{lrrrrrr}
\toprule
\multicolumn{7}{l}{\textit{A. Intact minus Reassigned: absolute-margin contributions}} \\
Objective & \multicolumn{2}{c}{Raw-correct} & \multicolumn{2}{c}{Raw-incorrect} & \multicolumn{2}{c}{Total} \\
\midrule
BT & \multicolumn{2}{c}{$+0.06935$} & \multicolumn{2}{c}{$+0.00965$} & \multicolumn{2}{c}{$+0.07900$} \\
NormBT & \multicolumn{2}{c}{$+0.05543$} & \multicolumn{2}{c}{$+0.00779$} & \multicolumn{2}{c}{$+0.06323$} \\
BSR & \multicolumn{2}{c}{$+0.08246$} & \multicolumn{2}{c}{$+0.01167$} & \multicolumn{2}{c}{$+0.09413$} \\
APLOT & \multicolumn{2}{c}{$+0.05854$} & \multicolumn{2}{c}{$+0.00923$} & \multicolumn{2}{c}{$+0.06776$} \\
DARM & \multicolumn{2}{c}{$+0.07028$} & \multicolumn{2}{c}{$+0.00800$} & \multicolumn{2}{c}{$+0.07829$} \\
\midrule
\multicolumn{7}{l}{\textit{B. Signed-margin contributions on the same fixed groups}} \\
& \multicolumn{3}{c}{Raw-correct} & \multicolumn{3}{c}{Raw-incorrect} \\
\cmidrule(lr){2-4}\cmidrule(lr){5-7}
Objective & Raw & Intact & Reassigned & Raw & Intact & Reassigned \\
\midrule
BT & $+0.8283$ & $+0.7316$ & $+0.6641$ & $-0.1341$ & $-0.1124$ & $-0.1048$ \\
NormBT & $+0.8251$ & $+0.7449$ & $+0.6915$ & $-0.1366$ & $-0.1127$ & $-0.1085$ \\
BSR & $+0.8239$ & $+0.7509$ & $+0.6704$ & $-0.1365$ & $-0.1103$ & $-0.1027$ \\
APLOT & $+0.8114$ & $+0.7094$ & $+0.6516$ & $-0.1459$ & $-0.1179$ & $-0.1096$ \\
DARM & $+0.8101$ & $+0.7216$ & $+0.6517$ & $-0.1469$ & $-0.1209$ & $-0.1133$ \\
\bottomrule
\end{tabular}
\par\vspace{4pt}
\begin{minipage}{.98\linewidth}\footnotesize Every entry averages all held-out prompts, including zero contributions outside the group, and is normalized by the held-out row-average raw margin magnitude. Panel A sums to the held-out magnitude difference. Panel B uses preferred-minus-less-preferred margins. This diagnostic uses 11,535 pairs, separately from the full-confirmation contrasts in Table~\ref{tab:objective-intervals}.\end{minipage}
\end{table}

\clearpage
\subsection{Independent temperature calibration}
For each objective, configuration, and seed, a positive temperature $T$ minimizes prompt-averaged negative log likelihood on the 11,026 calibration pairs. Evaluation uses disjoint prompts and preference probabilities $p_\theta=\sigma(d_\theta/T)$. With each pair ordered by its recorded preference, the losses are $-\log p_\theta$ and $(1-p_\theta)^2$. Table~\ref{tab:calibrated-predictions} reports these two losses and accuracy for all four configurations.

The Intact--Reassigned point estimates favor Intact for both losses under every objective. Their two-sided familywise intervals include zero. The table therefore presents probability quality alongside the magnitude comparison, while the main retention result describes the change in absolute clean margins. These paired loss comparisons use 50,000 shared-prompt and seed draws with a conservative family size of 20. Temperatures fitted on the independent calibration partition are held fixed during this loss inference. The numerical supplement contains the full loss differences and intervals.
\begin{table}[h]
\centering\tabfont
\setlength{\tabcolsep}{8pt}\renewcommand{\arraystretch}{1.14}
\caption{\textbf{Predictive losses after independent temperature calibration.} All four configurations use the same calibration and evaluation prompts within each objective and seed.}
\label{tab:calibrated-predictions}
\begin{tabular}{llrrr}
\toprule
Objective & Configuration & NLL & Brier & Accuracy (\%) \\
\midrule
BT & Raw & $0.51415$ & $0.17230$ & $73.60$ \\
 & Uniform & $0.51522$ & $0.17277$ & $73.52$ \\
 & Intact & $0.51349$ & $0.17215$ & $73.63$ \\
 & Reassigned & $0.51451$ & $0.17244$ & $73.62$ \\
\addlinespace[4pt]
NormBT & Raw & $0.51628$ & $0.17296$ & $73.47$ \\
 & Uniform & $0.51616$ & $0.17298$ & $73.51$ \\
 & Intact & $0.51551$ & $0.17279$ & $73.75$ \\
 & Reassigned & $0.51717$ & $0.17336$ & $73.35$ \\
\addlinespace[4pt]
BSR & Raw & $0.51742$ & $0.17348$ & $73.60$ \\
 & Uniform & $0.51530$ & $0.17273$ & $73.63$ \\
 & Intact & $0.51147$ & $0.17128$ & $73.82$ \\
 & Reassigned & $0.51500$ & $0.17270$ & $73.54$ \\
\addlinespace[4pt]
APLOT & Raw & $0.52928$ & $0.17851$ & $72.64$ \\
 & Uniform & $0.53156$ & $0.17912$ & $72.44$ \\
 & Intact & $0.52880$ & $0.17836$ & $72.64$ \\
 & Reassigned & $0.52978$ & $0.17846$ & $72.60$ \\
\addlinespace[4pt]
DARM & Raw & $0.53137$ & $0.17929$ & $72.42$ \\
 & Uniform & $0.53040$ & $0.17871$ & $72.60$ \\
 & Intact & $0.52977$ & $0.17841$ & $72.70$ \\
 & Reassigned & $0.53413$ & $0.18009$ & $72.31$ \\
\bottomrule
\end{tabular}
\par\vspace{4pt}
\begin{minipage}{.98\linewidth}\footnotesize A positive temperature is fit separately for each scorer on 11,026 calibration pairs; evaluation uses 11,535 disjoint pairs. Lower NLL and Brier values indicate better probability predictions. Prompt and seed averages follow Appendix~\ref{app:inference}. Complete paired intervals and absolute-metric intervals accompany the numerical data.\end{minipage}
\end{table}

\clearpage
\section{Prompt and Training-Seed Sensitivity}
\label{app:sampling-sensitivity}
The APLOT Uniform comparison separates two sources of sampling variation. Conditional prompt inference holds the three scorers at equal weight and samples their shared prompts. Crossed inference also samples the three seed weights independently of prompt weights. Calibration and evaluation prompts remain disjoint in both cases, and each draw fits the scaling coefficient on clean calibration prompts alone. This comparison preserves the endpoint populations and normalization of Appendices~\ref{app:scalar} and~\ref{app:aplot-formats}.

Table~\ref{tab:sampling-sensitivity} shows positive mean residuals and positive interval lower bounds under both schemes for aggregate edits, the three-format bank, and the nine-format mean. Crossed intervals are wider, reflecting variation among the observed scorers. Omitting any one seed also preserves the direction of all three average residuals. These results describe prompt sensitivity and consistency across the observed training repetitions; the empirical seed distribution has three support points.
\begin{table}[h]
\centering\tabfont
\setlength{\tabcolsep}{3pt}\renewcommand{\arraystretch}{1.14}
\caption{\textbf{Sampling sensitivity of APLOT Uniform extra attenuation.} Both sampling schemes use the same paired prompt draws and estimands.}
\label{tab:sampling-sensitivity}
\begin{tabular}{lrrrr}
\toprule
Endpoint & Mean & Fixed seeds & Crossed seeds & Leave-one range \\
\midrule
Aggregate & $0.0315$ & $[0.0252, 0.0380]$ & $[0.0067, 0.0662]$ & $[0.0170, 0.0409]$ \\
Three formats & $0.0191$ & $[0.0144, 0.0239]$ & $[0.0061, 0.0312]$ & $[0.0153, 0.0238]$ \\
Nine formats & $0.0111$ & $[0.0081, 0.0141]$ & $[0.0028, 0.0209]$ & $[0.0079, 0.0139]$ \\
\bottomrule
\end{tabular}
\par\vspace{4pt}
\begin{minipage}{.98\linewidth}\footnotesize Fixed-seed intervals condition on the three observed scorers; crossed intervals also resample their seed weights. Both use 50,000 draws, refit the clean scaling coefficient, and reestimate $s_0$. Two-sided familywise bounds retain $K=15$ for the aggregate and three-format endpoints and $K=30$ for the nine-format mean. Leave-one ranges contain the three two-seed means and are descriptive, not confidence intervals.\end{minipage}
\end{table}

\clearpage
\section{Anchor Strength and Sample Difficulty}
\label{app:anchor-difficulty}
The source constructor assigns anchor probability $a_i$ before mixing it into the soft target $t_i=0.9+0.1a_i$. To relate this correspondence to the learned scoring functions, we compare anchor strength with raw preferred-minus-less-preferred margins, their absolute magnitudes, and prediction correctness. This analysis uses the same 22,561 strict confirmation pairs and 12,554 prompts as the main profiles, with the eleven source strata used for assignment.

Within each source, fractional midranks describe anchor strength and each margin variable. We center these quantities by their prompt-weighted source means and correlate the centered values, weighting each pair by the inverse number of pairs in its prompt. Correctness uses the binary raw prediction indicator. The source adjustment separates within-source associations from differences between source means. Correlations are computed for each scorer and then averaged over training seeds.

Anchor strength is positively associated with signed margin, absolute margin, and correctness under all five objectives (Table~\ref{tab:anchor-associations}). For a complementary comparison, we fix two groups using the lower and upper halves of within-source anchor ranks, keeping tied values together. Higher-anchor pairs have greater raw accuracy and a larger Intact--Reassigned magnitude gain under every objective. Gains remain positive in the lower-anchor group as well. Thus the retained correspondence carries information related to sample difficulty within the evaluated representation family. The grouping and association analysis characterize this source construction; they do not establish transfer to independently produced preference labels.
\begin{table}[h]
\centering\tabfont
\setlength{\tabcolsep}{5pt}\renewcommand{\arraystretch}{1.14}
\caption{\textbf{Anchor strength and reward-model difficulty.} All five objectives use the same source-conditioned anchor ranks on 22,561 strict confirmation pairs.}
\label{tab:anchor-associations}
\begin{tabular}{lrrrr}
\toprule
\multicolumn{5}{l}{\textit{A. Source-adjusted associations with anchor strength}} \\
Objective & Signed margin & Absolute margin & \multicolumn{2}{c}{Correct prediction} \\
\midrule
BT & $0.586$ & $0.397$ & \multicolumn{2}{c}{$0.442$} \\
NormBT & $0.584$ & $0.395$ & \multicolumn{2}{c}{$0.435$} \\
BSR & $0.576$ & $0.389$ & \multicolumn{2}{c}{$0.431$} \\
APLOT & $0.562$ & $0.365$ & \multicolumn{2}{c}{$0.421$} \\
DARM & $0.562$ & $0.365$ & \multicolumn{2}{c}{$0.419$} \\
\midrule
\multicolumn{5}{l}{\textit{B. Common groups defined by within-source anchor rank}} \\
& \multicolumn{2}{c}{Raw accuracy (\%)} & \multicolumn{2}{c}{Intact--Reassigned magnitude gain} \\
\cmidrule(lr){2-3}\cmidrule(lr){4-5}
Objective & Lower & Upper & Lower & Upper \\
\midrule
BT & $57.22$ & $90.19$ & $0.0326$ & $0.1232$ \\
NormBT & $57.64$ & $89.86$ & $0.0220$ & $0.1059$ \\
BSR & $57.65$ & $89.77$ & $0.0428$ & $0.1465$ \\
APLOT & $56.83$ & $88.38$ & $0.0304$ & $0.1045$ \\
DARM & $56.56$ & $88.33$ & $0.0312$ & $0.1254$ \\
\bottomrule
\end{tabular}
\par\vspace{4pt}
\begin{minipage}{.98\linewidth}\footnotesize Panel A reports prompt-weighted, source-centered correlations. Anchor and margin variables use within-source fractional midranks; correctness is binary. Panel B fixes groups at anchor ranks $\leq0.5$ and $>0.5$, keeping tied anchors together. Group estimates preserve each pair\textquotesingle s full-population prompt weight and renormalize within the group. Gains use each seed\textquotesingle s main-analysis $s_0$. Entries average three seeds; all per-seed values and group weights accompany the numerical data.\end{minipage}
\end{table}

\end{document}